\documentclass{article}

\usepackage{arxiv}

\usepackage[utf8]{inputenc} 
\usepackage[T1]{fontenc}    
\usepackage{hyperref}       
\usepackage{url}            
\usepackage{booktabs}       
\usepackage{amsfonts}       
\usepackage{nicefrac}       
\usepackage{microtype}      
\usepackage{lipsum}		
\usepackage{graphicx}
\usepackage{doi}
\usepackage{amssymb}
\usepackage{amsmath}

\usepackage{apacite}

\title{Topological Data Analysis in Time Series: Temporal Filtration and Application to Single-Cell Genomics} 

\author{Baihan Lin\\
	Department of Systems Biology\\
	Columbia University Irving Medical Center\\
	New York, NY 10027 \\
	\texttt{baihan.lin@columbia.edu} \\
}

\renewcommand{\headeright}{}
\renewcommand{\undertitle}{Technical Report}

\hypersetup{
pdftitle={Topological Data Analysis in Time Series: Temporal Filtration and Application to Single-Cell Genomics},
pdfsubject={q-bio.NC, q-bio.QM},
pdfauthor={Baihan Lin},
pdfkeywords={First keyword, Second keyword, More},
}

\begin{document}
\maketitle

\begin{abstract}

The absence of a conventional association between the cell-cell cohabitation and its emergent dynamics into cliques during development has hindered our understanding of how cell populations proliferate, differentiate, and compete, i.e. the cell ecology. With the recent advancement of the single-cell RNA-sequencing (RNA-seq), we can potentially describe such a link by constructing network graphs that characterize the similarity of the gene expression profiles of the cell-specific transcriptional programs, and analyzing these graphs systematically using the summary statistics informed by the algebraic topology. We propose the single-cell topological simplicial analysis (scTSA). Applying this approach to the single-cell gene expression profiles from local networks of cells in different developmental stages with different outcomes reveals a previously unseen topology of cellular ecology. These networks contain an abundance of cliques of single-cell profiles bound into cavities that guide the emergence of more complicated habitation forms. We visualize these ecological patterns with topological simplicial architectures of these networks, compared with the null models. Benchmarked on the single-cell RNA-seq data of zebrafish embryogenesis spanning 38,731 cells, 25 cell types and 12 time steps, our approach highlights the gastrulation as the most critical stage, consistent with consensus in developmental biology. As a nonlinear, model-independent, and unsupervised framework, our approach can also be applied to tracing multi-scale cell lineage, identifying critical stages, or creating pseudo-time series.

\end{abstract}


\section{Introduction}

In recent years, technological developments in data visualizations, especially the subfield of topological data analysis (TDA), has illuminated the structure of biological data with features like clusters, holes, and skeletons across a range of scales \cite{carlsson2009topology}. The TDA approach has proven to be especially useful with recent advancements in experimental techniques at the single cell resolution, both in genomics and neuroscience, such as radiomics \cite{crawford2020predicting} and brain imaging \cite{saggar2018towards,phinyomark2017resting}. The utility of topology comes from the idea of persistence, which extract the underlying structures within data while discarding noisy elements in the single cell data collection. Unlike graph-based data like human connectomes, most of the time, the high-dimensional data collected from single cell techniques are similiarity-based. Under the assumption that these data was sampled from underlying space $\mathcal{X}$, the goal is to first approximate $\mathcal{X}$ with a combinatorial representation, and then compute some sort of invariant features to recover the topology of $\mathcal{X}$. \cite{amezquita2020shape,topaz2015topological,offroy2016topological} are a few recent reviews of the applications of TDA in various field of biology. \cite{chazal2017introduction} is a practical introduction and guide on how to apply TDA to data science and understand its results. \cite{otter2017roadmap} is a gentle introduction and tutorial to the computation
of persistent homology. 

The single-cell topological data analysis (scTDA) is one of the first attempts to apply topology-based computational analyses to study temporal, unbiased transcriptional regulation given the single-cell RNA sequencing data \cite{rizvi2017single}. In order to visualize the most invariant features of the entire gene expression data, scTDA clusters low-dispersion genes with significant gene connectivity according to their centroid in the topological representation, and visualize the data points in low-dimension space with the Mapper algorithm \cite{carlsson2014topological}. Computing the library complexity as the number of genes whose expression is detected in a cell, scTDA observes a mild dependence of library complexity over the timescale of the single cell data of 1,529 cells collected at 5 time points. This is expected because the number of genes expressed by cells in early stages of a developmental process is larger than in the adult case, as pointed out in \cite{gulati2020single}. As a result, in scTDA the library complexity is not used for any purpose at the topological data analysis and not related to any topological properties. 

Intuitively thinking, if we were to introduce a definition for ``cell complexity'', that characterizes the behaviors of cell-cell coexpression or interactions, the quantities of cell complexity should be agnostic to the number of genes expressed by the cells, and should be different across differentiated cells and across the developmental process. Can we introduce a better summary statistic for the cell complexity that can capture the developmental trajectory with more distinctions between time points? To clarify, unlike the previous definition of ``library complexity'', which simply quantifies the number of genes expressed in a cell, we wish to define a cell complexity measure to better model higher-order networks and dynamic interactions in single-cell data. Understanding the cell-cell interactions can help identify intercellular signaling pathways and previous analytical studies have focused on computing a communication score between the ligand–receptor pair of interacting proteins \cite{armingol2021deciphering}. For instance, \cite{arneson2018single} and \cite{oh2015extensive} infer the intercellular signaling pathways of cell-cell communications by computing the coexpression of all genes or other cell markers. The alternative would be to compute the similarity between gene expression profiles as in \cite{han2018mapping}. In this work, we aim to focus directly on the cell level, and use the similarity between each cell's gene expression profiles as a graph to compute a topological descriptor of the complexity. The more connected a group of cells are in this similarity graph, the higher the complexity of this group of cell is. There are two major quests in this line of research:

    \subsection{Quest from topological data analysis.}
Existing TDA applications usually focus on the low-dimensional graph visualization and the persistent homology of the data (i.e. computing the Betti numbers or barcodes up to dimension 2), because interpreting the biophysical meaning of the geometry and higher dimensional persistent modules is a conceptual challenge. Others have proposed hybrid approaches to combine the merits of data geometry and topology by adaptively selecting the proper thresholds in the pairwise distance matrix of the data points \cite{lin2018adgtic,lin2022geometric}. Another alternative to these low-dimensional TDA methods is the simplicial analysis. Simplicial architecture was studied in biological data through the application on human brain connectomes \cite{reimann2017cliques}, where each connected pairs of neurons are considered an edge to create a graph and the numbers of Rips-Vietoris simplices in dimensions up to 7 are computed at the static graphs comparing with the random graphs. Likewise in our inquiry, we are interested in the intercellular interaction within the same type of cells, the cell complexity, rather than the relationships between different groups of cell, as in scTDA \cite{simpArch}. However, the filtration challenge of deriving a graph from the distance-based data by choosing the best threshold, hinders the practical application of such simplicial analysis in these point cloud data. 

\subsection{Quest from single-cell-resolution data.}

With the increasingly popular usage of single-cell genomic techniques, it might be possible to infer such cell-cell interaction (or cellular ecology) in a fine resolution. However, as far as we are aware, there are only a few literature exploring the cellular ecology from single-cell RNA sequencing data. For instance, \cite{gallaher2020cells} and \cite{amend2016ecological} apply the ecology and multi-agent models to model single-cell systems. We wish to complement this line of work by connecting it to the topological data analysis, where the focus is to model the shape or manifold of the data from the similarity of data points. For instance, simplicial complexes are high dimensional objects or generalizations of neighboring graphs to represent the cliques of data points, and in other words, a notion of \textit{ecology}. The ecology doesn't have to be the organisms within a physical system. In the field of data science where we represent biological cells by their measurements (e.g. gene expression profiles) as data points residing in high-dimensional feature spaces, the ecology can be how these data points are connected to one another in the feature space. If we adopt an ecology research point of view, in order to characterize the dynamic systems of a community, one need to have knowledge or priors regarding the causal relationships between the agents (e.g. how do preys and predates interact, and in what ways). In order to parse out causal relationships, the temporal sequence of these events matters. Thus, the property of the synchrony and asynchrony of the events is a key to translate the feature space (represented by a similarity graph) to an ecology, which has directed (e.g. causal) relationships among the agents. This is why a temporal take into the topological data analysis can potentially unlock the first step from finding a static representations of the overall shape of the data points to discovering the event-directed representations (i.e. a temporal skeleton) of the data points.

One challenge of this hybrid direction, is to conceptually understand the biological meaning behind the dissimilarity of the omic data. For instance, what does it mean if two cells have similar gene expression profiles from each other? Does that indicate a homogeneity if the two cells are from the same tissues, or is it an artifact that the manual labeling or classifications are not perfect? Can we measure the ``complexity'' of the cell populations based on the heterogeneity or diversity within populations? If we can, how to we evaluate and interpret lower-order versus higher-order ``complexity''? These are some open questions we wish to engage the field to discuss and investigate together, instead of answering them directly in this first work. 

The other challenge is the scalability and compariablity of the single-cell data. With the advances of multi-channel high-throughput data collection techniques in biological fields, how to compute the pairwise distances of the point clouds efficiently? In different trials of single-cell experiments, how to make sure that the persistent modules are comparable to one another? 

\begin{figure}[tb]

\includegraphics[width=\linewidth]{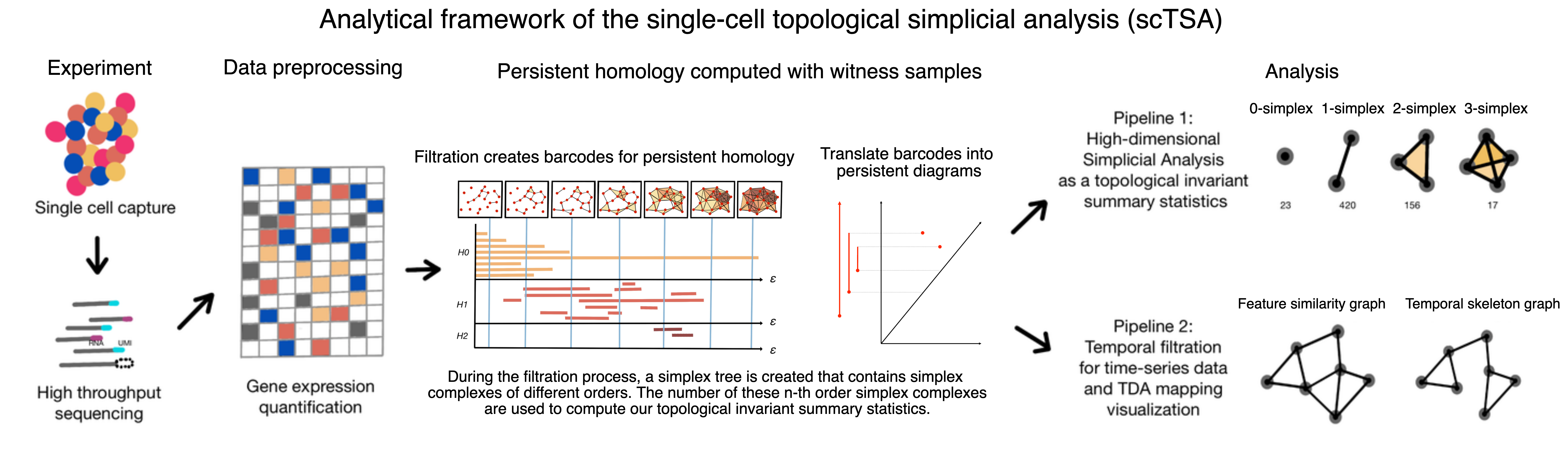} \hfill
\caption{\textbf{The analytical framework of the single-cell topological simplicial analysis (scTSA)}. The pipeline starts with the single-cell sequencing data, which is then preprocessed into gene expression profiles in a 2d matrix (rows are cells and columns are genes). This step can also go through another layer of dimension reduction. From the matrix, we compute filtrations over their feature space and temporal constraints. The persistent homology can be computed from the filtration process with either a persistent barcode or a persistent diagram. The filtrations obtained through the processes can also used for simplicial analysis, which groups the cells by time steps before the analysis. Finally, one can visualize the data using the Mapper algorithm, with or without the temporal constraints.}
\label{fig:pipeline}

\end{figure}

\subsection{Framework: single-cell topological simplicial analysis (scTSA)}

In this study, we propose a topological simplicial analysis (TSA) pipeline (Figure \ref{fig:pipeline}) as an exploratory inquiry to solve these three challenges: (1) with the algebraic geometry's definitions of forming higher-order simplices, we can potentially interpret that cliques of higher orders indicates operational units of higher order; (2) with the bootstrapping techniques to sample from the data points collected at each sub-level, we can scale the analysis to large single cell datasets and compare groups of cells quantitatively; (3) with a time delay constraint on the filtration process, we can sort the projected data points of cells into distinct groups of cells collected from the same time stamps. The framework first takes the measurements of the single-cell RNA sequencing data which generates a similarity matrix among the cells based on their gene expression profiles. Other than performing the persistent homology to obtain lower-order topological descriptors of the data, we compute additional higher-order topological descriptors by counting the number of the simplices emerged from the filtration process. In addition, we introduce a technique to extract the temporal skeleton of the developmental processes, called temporally filtrated TDA, and show that the developmental trajectories of cells can be better revealed in this approach comparing to existing TDA mapping techniques. 

We begin our presentation in section \ref{sec:method}, with a short overview of mathematical definitions of the single cell data visualization problem and introduction of necessary concepts and definitions in the language of computational topology. Section \ref{sec:method} formulates the topological simplicial analysis pipeline we are proposing as well as numerical tricks applied in the implementation to ensure the scalablity. We apply this single cell Topological Simplicial Analysis (scTSA) to the zebrafish single-cell RNA sequencing data with 38,731 cells, 25 cell types, over 12 time steps \cite{farrell2018single}. We select the top 103 genes based on the scTDA pipeline from the high-dimensional high-throughput transcriptomic data. In section \ref{sec:results}, we introduce the dataset used to benchmark the method and present the analysis results with their mathematical interpretations to the biological insights. In the last section, we discuss the validity of using our framework to understand the higher-order cellular complexity, and conclude our methods by pointing out several future work directions as the next step of this line of research.

\section{Materials and Methods}
\label{sec:method}

\begin{figure}[tb]

\includegraphics[width=\linewidth]{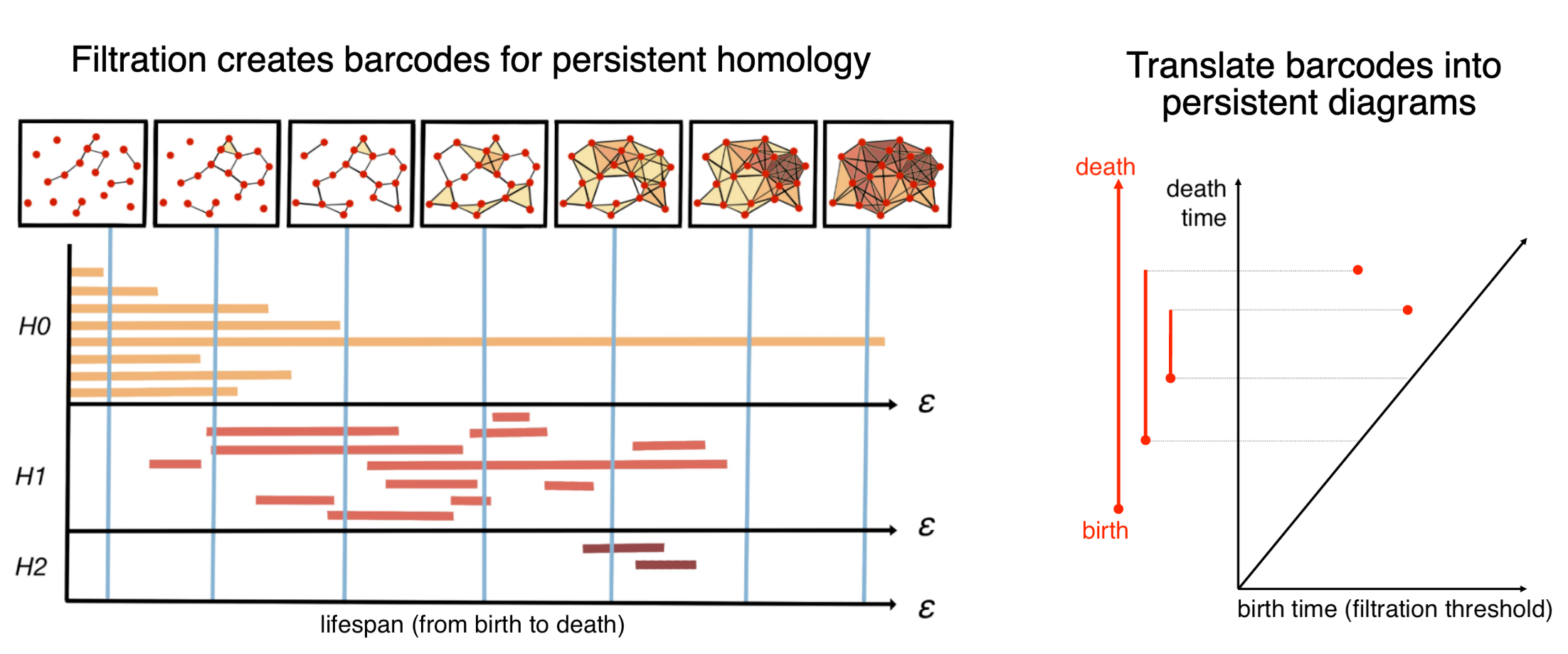} 
\caption{
\textbf{Persistent homology via mathematical filtration.} In this schematic diagram, a point cloud of 19 data points are presented in a low-dimensional embedding space. In the filtration process, a parameter $\epsilon$ is swept from 0 to the maximum pairwise distance within the point cloud, indicating a distance threshold under which the two points can form an edge to become one connected component in the graph. For each value $\epsilon$, we obtain a space $S_\epsilon$ consisting of vertices, edges formed among the vertices, and higher-dimensional polytopes connected by these edges. For instance, a nerve ball of radius $\epsilon$ grows around each point cloud, and an edge will form if two nerve balls touch. $H_n$ indicates the $n$-th homology group, i.e. the formation of the simplex complexes of order $n$, with 0-simplex to be the nodes (or clusters), 1-simplex to be the edges between two nodes, 2-simplex to be the loops (or triangles in this case), 3-simplex to be the tetrahedrons and so on. We log the existence of a n-simplex if and only if all of its components (e.g. (n-1)-simplex, (n-2)-simplex, ...1-simplex, and 0-simplex) are all in $S_\epsilon$. Each colored line indicates the ``lifespan'' of a simplex, with its starting point to be its ``birth'' (or first appearance) and ending point to be its ``death'' (or disappearance due to the two nerve balls fully overlapping). In this example, the persistent homology of the data cloud can be presented in the form of a ``barcode'' representation, which is a finite collection of intervals. The birth and death of the simplicial complexes up to the order 2 are recorded when the filtration process gradually sweeps the distance threshold. The barcode representation is often replaced with the visualization of a 2d persistent diagram, in which the x-axis indicates the birth time (the distance threshold a filtration appears) and the y-axis indicates its death time (the distance threshold the filtration disappears).}
\label{fig:PH}

\end{figure}

\begin{figure}[tb]

\includegraphics[width=\linewidth]{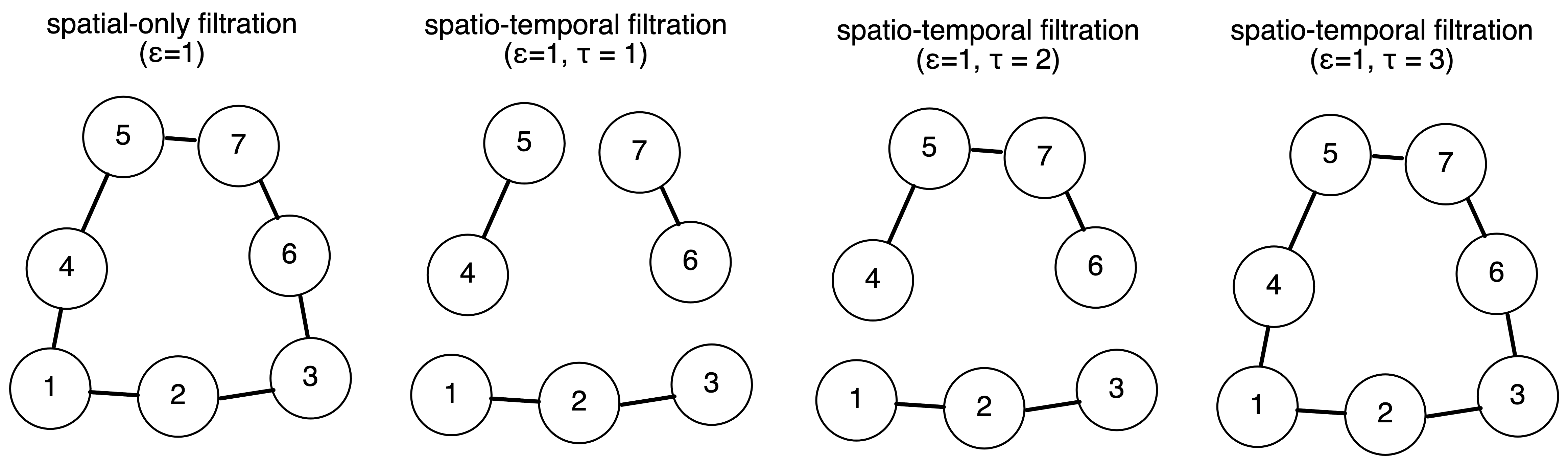} \hfill
\caption{\textbf{Intuitive example of the temporal filtration}. Presented here are 7 data points, which are marked by their time stamps (when they are measured). In all four cases, we consider the case where the spatial threshold $\epsilon$ of the nerve ball around each data point is 1 (which in our case only contains every data point's nearest neighbor, but not their second nearest neighbor). If we only perform the spatial filtration, we would consider them all to be connected. However, that would not match the temporal skeleton. Instead, we can set a temporal constraint $\tau$ such that only if two data points that are spatially (in the feature space) proximal to each other are also measured temporally close to each other, their edge is included. If $\tau$ is small (say, 1 time step apart), we have a fine resolution temporal skeleton which separate the data points into three main phases. If $\tau$ is big (say, 3 time steps part), we have a crude temporal skeleton which groups them all in a connected components.
}
\label{fig:tfilter}

\end{figure}

\begin{figure}[tb]

\includegraphics[width=\linewidth]{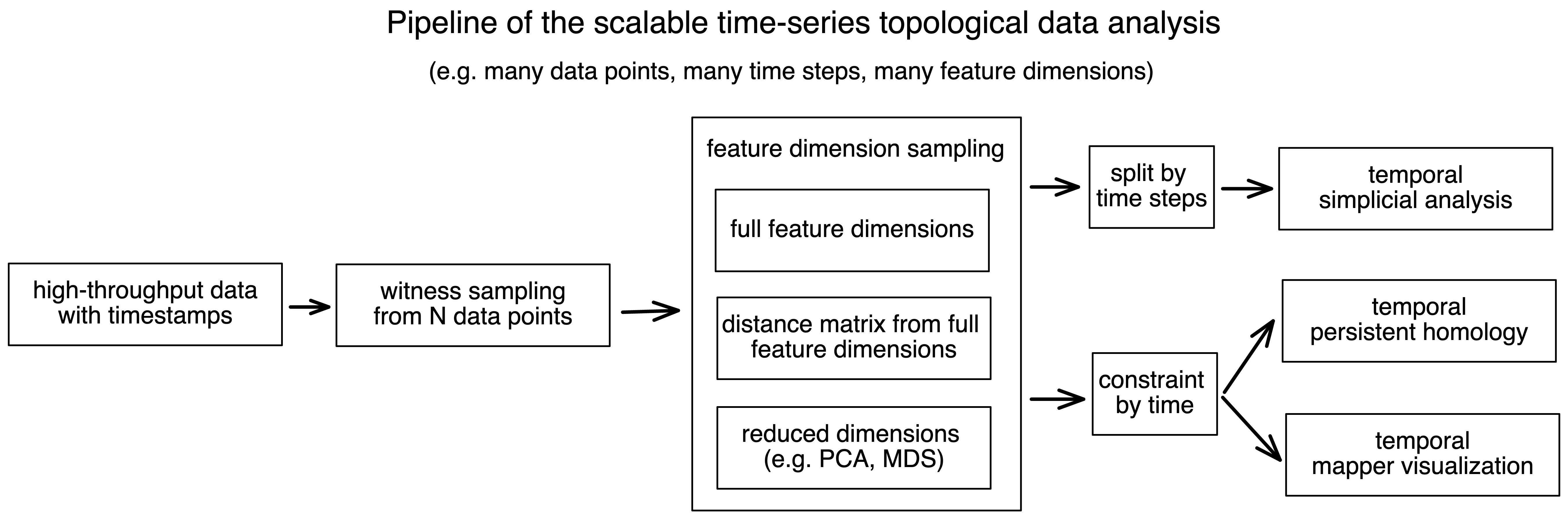} \hfill
\caption{\textbf{Pipeline of the time-series topological data analysis for high-throughput data}. We start with the high-throughput data points marked with their timestamps. To decrease the number of data points for efficient computation (and also comparability across time points), a witness sampling is performed among these data points. Then one can choose to reduce the dimension or not given the noise and distribution properties of their data. To perform the temporal simplicial analysis, the data points are first separately grouped into different time points, and then computed their filtrations to obtain their number of simplicial complexes at different orders. To perform the temporal persistent homology and mapper visualization, one can apply the temporal constraint onto the sampled data points so far to obtain a temporal skeleton.}
\label{fig:scalable}

\end{figure}

\subsection{Single-cell data in the point cloud space}

Genomic measurement and analysis at single-cell resolution has enabled new understandings of complex 
biological phenomena, such as revealing cellular composition of complex tissues and organisms \cite{kalisky2011single}. Single-cell RNA sequencing (scRNA-seq) techniques measure the gene expression profiles of individual cells through mechanisms like microfludics. For instance, the benchmark dataset of zebrafish embryogenesis \cite{farrell2018single} that we use in this study, applied Drop-seq, a massively parallel scRNA-seq method to profile the transcriptomes of tens of thousands of embryonic cells \cite{macosko2015highly}. These single cell data are usually point clouds in a finite metric space, a finite point set $S \subseteq {\mathbb R}^d$. Let $d(\cdot, \cdot)$ denote the distance between two points in metric space $\mathcal{Z}$. The assumption is that data was sampled from underlying space $\mathcal{X}$. The goal is to recover topology of $\mathcal{X}$. To accomplish the goal, one needs to first approximate X with a combinatorial representation (e.g. with the simplicial complex), and then compute a topological invariant summary statistics (e.g. with the persistent homology).

\subsection{Definition of the simplicial and temporal filtration}
\label{sec:temporal_filtration}

Given the point cloud data, we then construct a continuous shape on top of the data to highlight the underlying topology and geometry. The process to build such a shape is through a mathematical filtration, which is often a simplicial complex or a nested family of simplicial complexes, that reflects the innate structure of the point cloud data at different scales \cite{chazal2017introduction}. 
If we consider all the points in the point cloud data each with a coordinate of their locations in certain embedding, they each occupy a spherical space with the same radius $\epsilon$ around them, which are called nerve balls. If the two nerve balls overlap or contact each other, we consider an edge to be formed between them in this graph. The filtration is a process to tune the parameter $\epsilon$ from $0$ to $\infty$ and record the families of simplicial complexes generated through the increasingly connected (or ``complex'') graph.

Usually, the challenge is to extract relevant and useful information about the shape of the data through defining such simplicial complexes from the graph (generated through the filtration process). Rips-Vietoris complex is one of the common choices in practice to compute topological invariants of point clouds, defined as follows: given the vertex set $\mathcal{Z}$, for each pair of vertices $a$ and $b$ edge a-b is included in Rips-Vietoris complex $C(\mathcal{Z}, t)$ if $d(a, b) \leq t$, and a higher dimensional simplex is included in $C(\mathcal{Z}, t)$ if all of its edges are included. Since $C(\mathcal{Z}, t) \in C(\mathcal{Z}, t')$ whenever $t \leq t'$, the filtered Rips-Vietoris complex is a filtered simplicial complex, and also the maximal simplicial complex that can be built on top of its $1-$skeleton, thus a clique complex or a flag complex.
Unlike conventional low-dimensional topological data analysis, we computed simplices into high dimension (up to 7) during the entire filtration process. To record the number of cliques, we compute the filtered simplicial complexes and record their cumulative counts across the entire filtration process. 

Since the topological data analysis usually only consider the graph constructed by the spatial proximity (i.e. the distance matrix) between the data points in the low-dimensional embedding, it is not clear how to incorporate timestamp information for meaningful inference and visualization when facing the time-series data streams. One approach would be to simply consider the time stamp as the meta data for post hoc labeling of the topological representations. Another alternative would be to consider time as an additional dimension in the filtration process. We present the Temporal Filtration as the following: alongside the conventional sweeping of the parameter $\epsilon$ from $0$ to $\infty$, we set another parameter $\tau$ to indicate a hard constraint in edge forming between two points. Alternatively and intuitively, the temporal filtration is equivalent to conventional filtration by using the composite norm: 

\begin{equation}\label{eq:composite}
d((x,t_x), (y,t_y)) = max( \frac{1}{\epsilon^*} |y-x|, \frac{1}{\tau^*} |t_y-t_x| )
\end{equation}

where the $\epsilon^*$ and $\tau^*$ are directly relates to the spatial threshold $\epsilon$ (in the feature space) and the temporal threshold $\tau$.
As a practical note from this notation, it can be used without additional specialist software.

in other words, only if the time stamp difference between the two data points is within the time delay limit $\tau$, can two nerve balls, if spatially proximal enough (less than $\epsilon$), form an edge in between. On the other hand, if the time stamp difference between the two data points is larger than $\tau$, even if they are spatially proximal enough (less than $\epsilon$), they cannot form an edge. Given the problem settings, one can either set a reasonable time delay limit $\tau$ given the domain knowledge, or tune $\tau$ from $0$ to $\infty$, similar to the filtration process on the spatial filtration parameter $\epsilon$. The later approach can potentially extract temporally invariant topological summary statistics.

Figure \ref{fig:tfilter} is an intuitive example of the criterion of edge forming in the temporal filtration. In the example, we have 7 data points, which are marked by their time stamps (when they are measured). The node marked 1 would indicates it is collected the time step 1. There is a 1 time step difference between each data point of consecutive numbers. To illustrate the differences between the conventional and temporal filtration, the schematic is a snapshot of the full filtration process, frozen at a set of filtration thresholds.
In all four cases, we consider the case where the spatial threshold $\epsilon$ of the nerve ball around each data point is 1 (which in our case only contains every data point's nearest neighbor, but not their second nearest neighbor). If we only perform the spatial filtration, we would consider them all to be connected. However, that would not match the temporal skeleton. Instead, we can set a temporal constraint $\tau$ such that only if two data points that are spatially (in the feature space) proximal to each other are also measured temporally close to each other, their edge is included. If $\tau$ is small (say, 1 time step apart), we have a fine resolution temporal skeleton which separate the data points into three main phases. If $\tau$ is medium (say, 2 time step apart), we have a relatively crude resolution temporal skeleton which separate the data points into two main phases. If $\tau$ is big (say, 3 time steps part), we have the crudes temporal skeleton which groups them all in a connected components. This also demonstrates the possibility of using $\tau$ as a hierarchical mechanism to parse persistent features of different temporal resolutions.

\subsection{Topological data analysis with persistent homology}

Following the definition above, an abstract simplicial complex is given by a set $\mathcal{Z}$ of vertices or $0-$simplices, for each $k \leq 1$ a set of $k-$simplices $\sigma = [z_0,z_1,\dots,z_k]$ where $z_i \in \mathcal{Z}$, and for each $k-$simplex a set of $k + 1$ faces obtained by deleting one of the vertices. A filtered simplicial complex is given by the filtration on a simplicial complex $\mathcal{Y}$, a collection of subcomplexes $\{\mathcal{Y}(t)|t\in {\mathbb R}\}$ of $\mathcal{Y}$ such that $\mathcal{Y}(t) \subset \mathcal{Y}(t')$ whenever $t\leq t'$. The filtration value of a simplex $\sigma\in\mathcal{Y}$ is the smallest $t$ such that $\sigma\in\mathcal{Y}(t)$. 
Topological data analysis methods usually involve computing the persistent homology \cite{de2004topological}. The Betti numbers help describe the homology of a simplicial complex $\mathcal{Y}$. The Betti number value $BN_k$, where $k \in \mathbb{N}$, is equal to the rank of the $k-$th homology group of $\mathcal{Y}$. 
The Betti intervals over the filtration process help describe how the homology of $\mathcal{Y}(t)$ changes with $t$. A $k-$dimensional Betti interval, with endpoints $[t_{\text{start}}, t_{\text{end}})$, corresponds to a $k-$dimensional hole that appears at filtration value $t_{\text{start}}$, remains open for $t_{\text{start}} \leq t < t_{\text{end}}$, and closes at value $t_{\text{end}}$. 

Figure \ref{fig:PH} is a schematic diagram outlining how to perform a filtration process (by sweeping the $\epsilon$), document the ``birth'' and ``death'' of each complexes (the colored lines of various length in the chart), and generate this as a barcode representation \cite{ghrist2008barcodes} or a persistent diagram \cite{cohen2005stability} for the downstream analyses. In this schematic diagram, a point cloud of 19 data points are presented in a low-dimensional embedding space. 
In the filtration process, a parameter $\epsilon$ is swept from 0 to the maximum pairwise distance within the point cloud, indicating a distance threshold under which the two points can form an edge to become one connected component in the graph. For each value $\epsilon$, we obtain a space $S_\epsilon$ consisting of vertices, edges formed among the vertices, and higher-dimensional polytopes connected by these edges. For instance, a nerve ball of radius $\epsilon$ grows around each point cloud, and an edge will form if two nerve balls touch. $H_n$ indicates the $n$-th homology group, i.e. the formation of the simplex complexes of order $n$, with 0-simplex to be the nodes (or clusters), 1-simplex to be the edges between two nodes, 2-simplex to be the loops (or triangles in this case), 3-simplex to be the tetrahedrons and so on. We log the existence of a n-simplex if and only if all of its components (e.g. (n-1)-simplex, (n-2)-simplex, ...1-simplex, and 0-simplex) are all in $S_\epsilon$. Each colored line indicates the ``lifespan'' of a simplex, with its starting point to be its ``birth'' (or first appearance) and ending point to be its ``death'' (or disappearance due to the two nerve balls fully overlapping). 
In this example, the persistent homology of the data cloud can be presented in the form of a ``barcode'' representation, which is a finite collection of intervals. The birth and death of the simplicial complexes up to the order 2 are recorded when the filtration process gradually sweeps the distance threshold. The barcode representation is often replaced with the visualization of a 2d persistent diagram \cite{cohen2005stability}, in which the x-axis indicates the birth time (the distance threshold a filtration appears) and the y-axis indicates its death time (the distance threshold the filtration disappears). In most cases, only the first two orders of the filtrations are computed and included in persistent barcodes or diagrams.

By using the temporal filtration in place of the conventional filtration, we can extend the methods of persistent homology into one for temporal persistent homology. Our method is related to the research of multi-parameter persistence \cite{botnan2022introduction}, which aims to construct a topological space with more than one filtered spaces.
In other words, the computation of a persistent barcode or diagram can also be customized to use temporal filtration as its filtration criterion, by either using a composite norm function as in Eq. \ref{eq:composite}, or using a multi-parameter filtration with a temporal constraint. 

\subsection{Empirical simplicial computation with witness sampling and dimension reduction}

Overall, the witness sampling is critical for two reasons: (1) The single cell data has different noise granularity across cell types and data collection procedures \cite{faure2017systematic}, and thus, the number of cells collected in each time points and different cell types (as in the analyzed developmental study \cite{farrell2018single}) can vary in different magnitude, making direct simplicial computation incomparable. (2) In large-scale high-throughput data, the large number of data points and feature sizes can make computation especially expensive and infeasible. For instance, the computation of filtration requires a comparison between a sweeping proximity threshold and the distance between two data points, and computing the distance matrix between all points is not only time-consuming and memory-exhaustive (e.g. 1M points would be 1T to just storing the distance matrix).

For these larger datasets, if we include every data point as a vertex, the filtrated simplicial complexes can quickly contain too many simplices for efficient computation. To solve this numerical inconsistency issue, we instead extract the lazy witness complexes by sampling $m$ data points \cite{de2004topological} with a sequential maxmin procedure \cite{adams2009nonlinear}, setting a nearest neighbor inclusion of 2 (as in the term ``lazy'')\footnote{The selection of $m$ depends on the scale of the dataset. The bigger the sample size $m$ is, the better the estimate. However, since different partitions of the data points have varying sizes. For instance, if there are only 50 data points collected in time step 1, while there are more than 100 points in other times steps, then the maximum of $m$ that can be picked is 50.}. The computation of the witness complex in high dimensions can be implemented with GUDHI \cite{maria2014gudhi}, Ripser \cite{bauer2021ripser}, and JPlex software \cite{sexton_jplex_2008}. 
The codes to reproduce the empirical results can be accessed at \href{https://github.com/doerlbh/scTSA}{\underline{https://github.com/doerlbh/scTSA}}. 

Figure \ref{fig:scalable} outlines our scalable time-series topological simplicial analysis pipeline. We start with the high-throughput data points marked with their timestamps. To decrease the number of data points for efficient computation (and also comparability across time points), a witness sampling is performed among these data points. Then one can choose to reduce the dimension or not given the noise and distribution properties of their data. The usage of dimension reduction is a useful step before the filtration. Due to the ``Curse of Dimensionality'', the data points in a very high-dimensional space can be very sparse and thus the distances between them usually collapses to a constant, i.e. residing at a hyperspherical space. As a result, the filtration computation around them can be ineffective and unstable. Mapping them onto a low-dimensional space can partly solve this issue.

Then to perform the temporal simplicial analysis, the data points are first separately grouped into different time points, and then computed their filtrations to obtain their number of simplicial complexes at different orders. To perform the temporal persistent homology and mapper visualization, one can apply the temporal constraint onto the sampled data points so far to obtain a temporal skeleton.

\subsection{Topological simplicial analysis}

Given the simplicial complexes of different orders from the witness sampling approach, we need to correct for the effect of sampling. The larger the sample size, the more likely the higher-order simplicial complexes emerge. One way to correct for this amplification effect is to normalize this quantity directly to the quantity collected from a null distribution of the data.
Usually for a graph, network or more generically, data with a binary connectivity format (e.g. brain connectome), the Erd{\H{o}}s-R{\'e}nyi random graph \cite{erdHos1960evolution} can be used as control models. However, in fully connected similarity-based data, the average connectivity probability is entirely dependent on the filtration factor. To avoid this caveat, we take a different approach by permuting the pairwise distances of the data points, which is equivalent to a weighted version of the Erd{\H{o}}s-R{\'e}nyi random graph. Another strategy would be permuting the feature at each dimension. In this way, the low-dimensional embeddings computed by the multidimensional scaling can form different connectivity profiles while maintaining the same distance distribution. Then we apply the same topological data analysis pipelines to the embeddings computed from the pairwise distance matrices from both the actual data and the control models. 

To this point, we propose a formal definition of cellular complexity, as the \textit{normalized n-simplicial complexity}, $NSC_n$, a family of summary statistics with an increasing order $n$:

\begin{equation}
    NSC_n = \frac{SC_n^{data}}{SC_n^{null}}
\end{equation} 

where $NSC_n$ is computed by taking the ratio between the number of the simplicial complexes for a certain order $n$ computed from the actual data (which we denote $SC_n$), to the sum of the number of those computed from the control models and from the actual data. An alterantive would be $NSC_n = \frac{SC_n^{data}}{(SC_n^{data}+SC_n^{null})}$. A value of 0.5 would indicates that the simplicial complexity at order n is the same in the data and the null models. Empirically, we compute the $NSC_n$ with the order \textit{n} from 1 to 7, as the summary statistics characterizing the ecology among the data points with cliques and cavities of increasing modularities.

\subsection{Topological data visualization with low-dimensional mapping}

To build and visualize the topological representation of the point cloud data, we use the Mapper algorithm \cite{singh2007topological} through the implementations provided by Kepler-Mapper \footnote{https://github.com/scikit-tda/kepler-mapper} with modifications for temporal filtration
at \href{https://github.com/doerlbh/tkMapper}{\underline{https://github.com/doerlbh/tkMapper}}. 
In brief, a dissimiliarity matrix is computed from the preprocessed RNA-seq data by taking the pairwise correlation distance. This metric space was then reduced to a low-dimensional embeddings with the multi-dimensional scaling \cite{mead1992review}. Given this embedding, the point cloud data are chopped into coverings of hypercubes with a 50\% percentage of overlapping between the cubes\footnote{The choice of 50\% is empirically determined by our dataset. We vary the overlap parameter among 25\%, 50\% and 75\%, and 50\% gives the best clustering effect.}. Then for each hypercube, the data points within the cube are then clustered with single-linkage rule. This step further aggregates all the points into a network in which each vertex corresponds to a cluster and each edge corresponds to a non-vanishing intersection between the clusters. As defined in section \ref{sec:temporal_filtration}, if temporal filtration is applied, then edge forming is also controlled by the additional time delay constraint that the clusters are formed with both spatial and temporal proximity, and the edges would only exist between two clusters if all points in the two clusters are within the time delay limit $\tau$. In other words, the filter function is the same that we apply to persistent homology, which can either be a single filtration with the temporal constraint, a single filtration with the temporal composite norm, or a multi-parameter filtration.
Once we reach a network representation, the network can eventually be visualized with force-directed algorithms for insights.


\section{Results}
\label{sec:results}

\begin{figure}[tb]

\includegraphics[width=\linewidth]{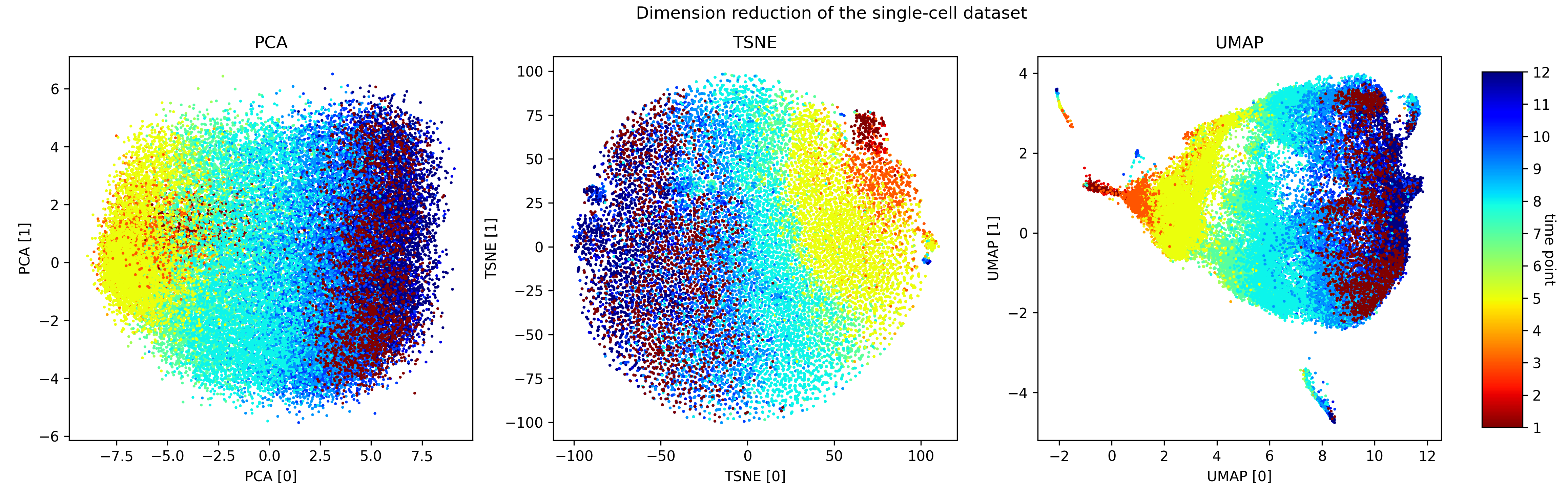} \hfill
\caption{\textbf{Dimension reduction of the dataset}. PCA, TSNE and UMAP applied to our preprocessed gene expression profiles.}
\label{fig:dr}

\end{figure}

\begin{figure}[tb]

\includegraphics[width=\linewidth]{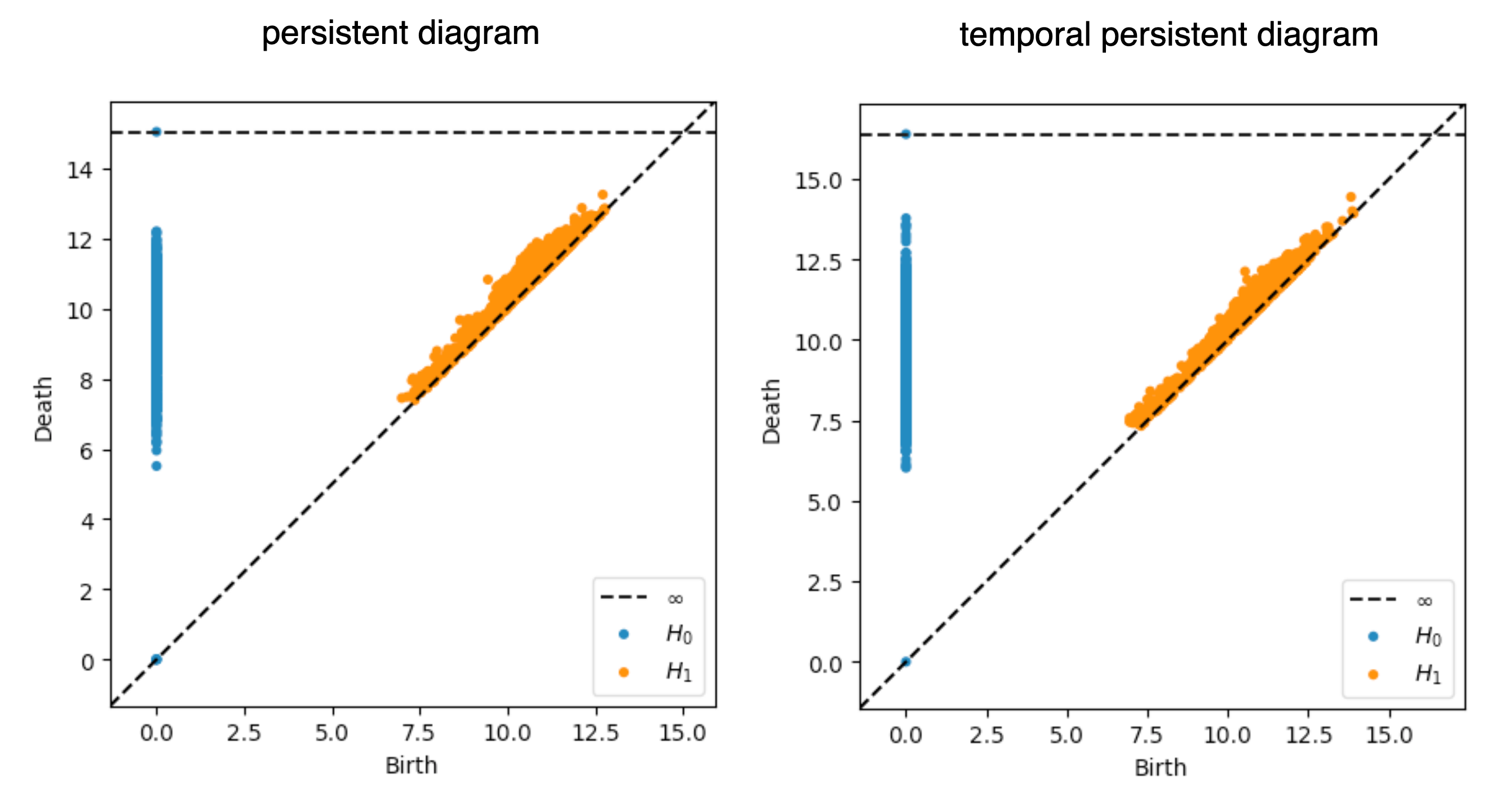} \hfill
\caption{\textbf{Persistent diagrams}. The persistent diagrams computed from the persistent homology and temporal persistent homology is shown here. The x-axis corresponds to the birth of all the persistent modules arises in the filtration process, and the y-axis corresponds to their death.}
\label{fig:tph}

\end{figure}

\begin{figure*}[tb]

\centering
\includegraphics[width=\linewidth]{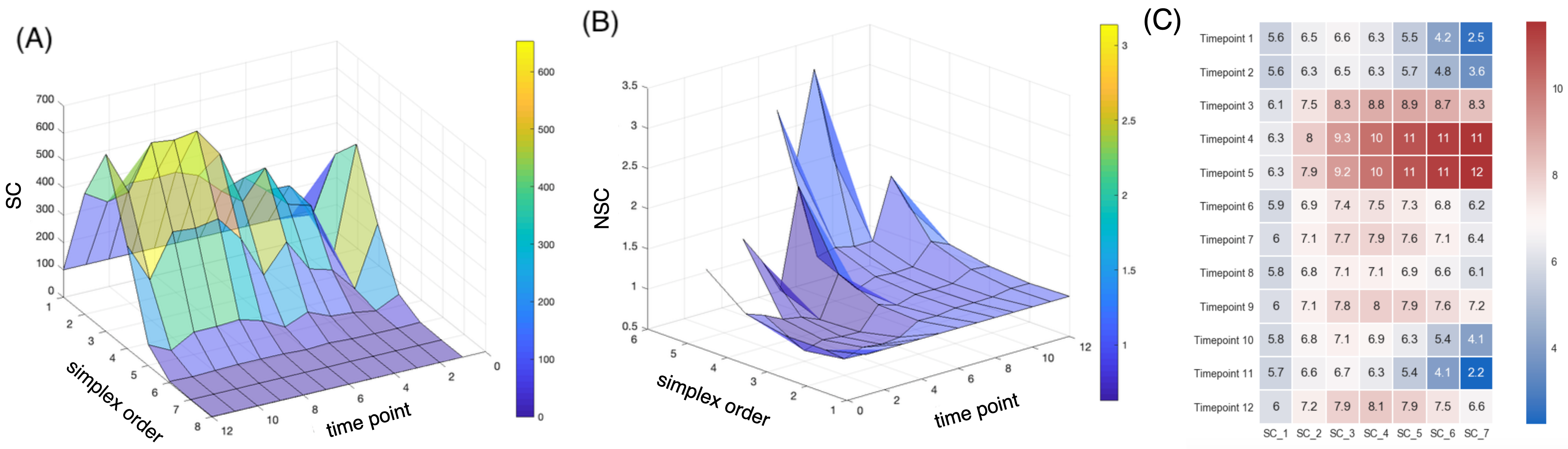}
\caption{\textbf{Simplicial dynamics across developmental stages.} (A) The number of $n$-simplices is computed from the sampled data points in each time points. (B) The normalized $n$-simplicial complexity, i.e. the normalized number of $n$-simplices, is computed as the ratio of the number of the $n$-th order simplicial complexes from the data over the number of those from the null models. The normalized simplicial complexity of higher order appears to be well above 1 in certain developmental stages with a distinctive separation between the 5th and 6th time points. (C) The heatmap of the normalized $n$-simplicial complexity across the time points supports the observation. 
}
\label{fig2}

\end{figure*}

\begin{figure*}[tb]

\centering
\includegraphics[width=0.48\linewidth]{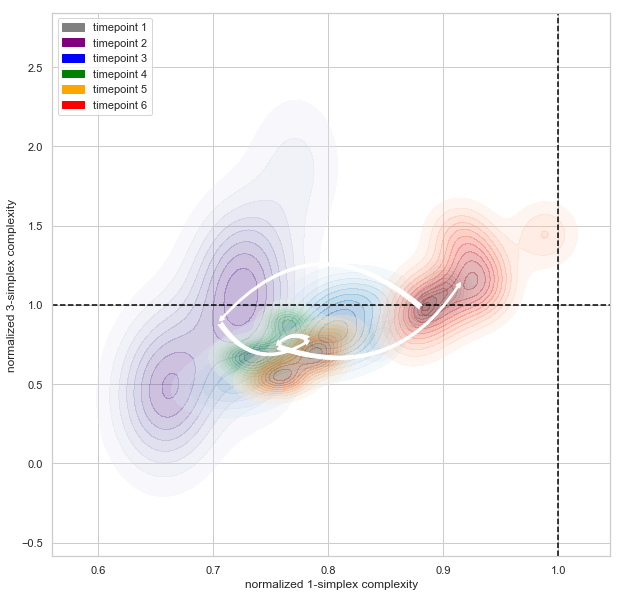}
\includegraphics[width=0.48\linewidth]{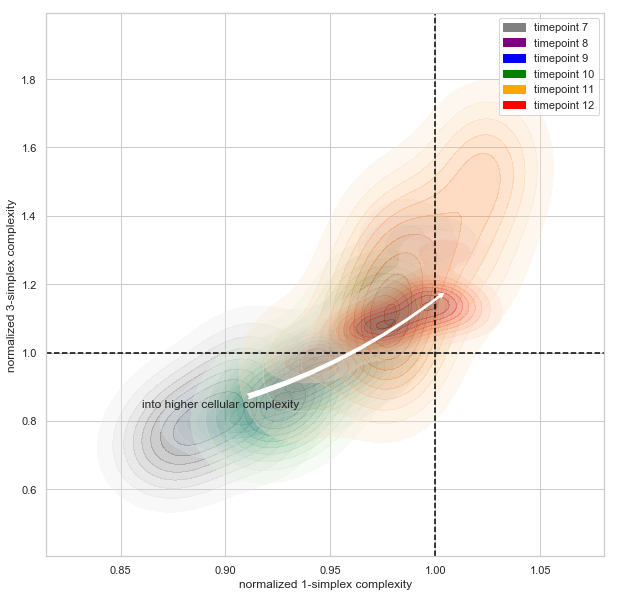}
\caption{\textbf{Simplicial dynamics across developmental stages.} To investigate the tradeoff between the higher-order and the lower-order simplicial complexity in the developmental stages, the normalized 3-simplicial complexity is mapped against the normalized 1-simplicial complexity. The color indicates different time points. The arrow indicates the transition between the centroids in each groups of time points. A transition of lower-order and higher-order normalized cell complexity is marked with the white trajectories across sequential time points.}
\label{fig2C}

\end{figure*}

%
%

\begin{figure*}[tb]

\includegraphics[width=1\linewidth]{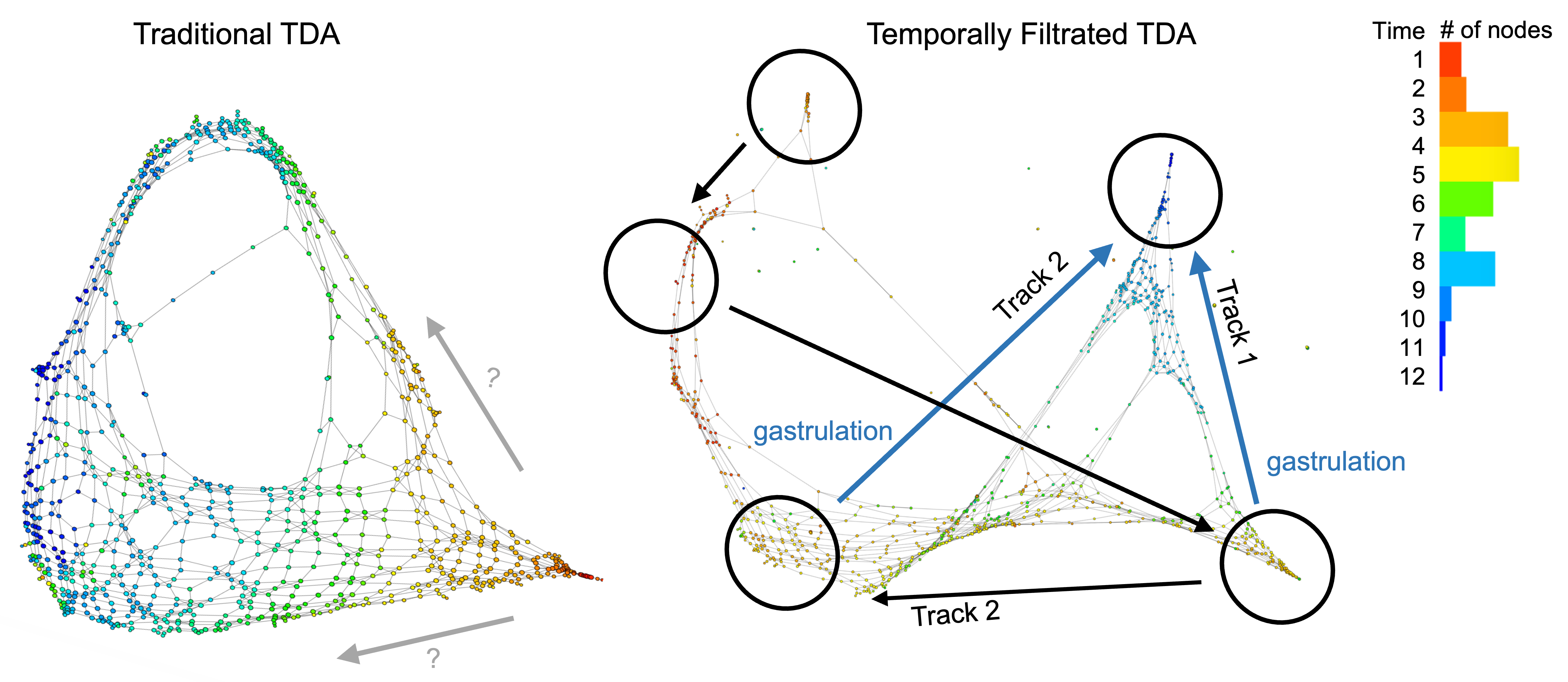}
\caption{\textbf{Temporal filtration identifies the critical stage of cellular complexity change.} The color indicates the time points and each node corresponds to a small cluster of cells collected at the same time points. The conventional TDA mapping (the left panel) identifies a bifuraction structure, but there are spatial locations that has a mixture of clusters that belong to non-consecutive time points. This makes the identifications of a developmental pathway challenging. When applying the temporal filtration (the right panel), the mapping identifies a cleaning separation of two tracks, or two subpopulations of cells that evolves in the gastrulation stage, matching the observation in our summary statistics from the algebraic topology.}
\label{fig3C}

\end{figure*}

\begin{figure*}[tbh]

\includegraphics[width=1\linewidth]{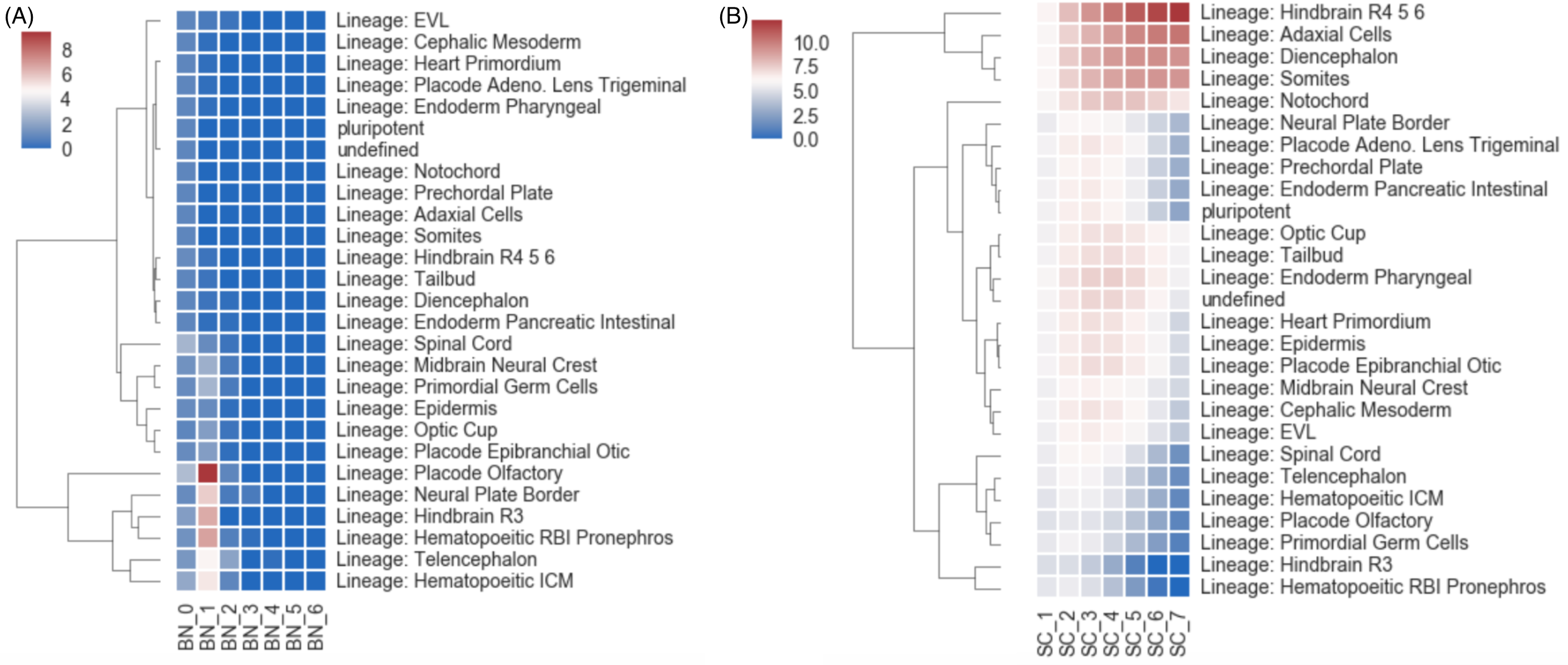}
\caption{\textbf{Cell lineage tracing with the simplicial statistics.} In this analysis, the hierarchical clustering is performed on the summary statistics of transcriptomic data of different cell types. (A) The heatmap and clustering result using the Betti numbers as the clustering features. (B) The heatmap and clustering result using the normalized simplicial complexity as the features for the hierarchical clustering.}
\label{fig:lineage}

\end{figure*}

\begin{figure}[tb]
\centering
\includegraphics[width=.8\linewidth]{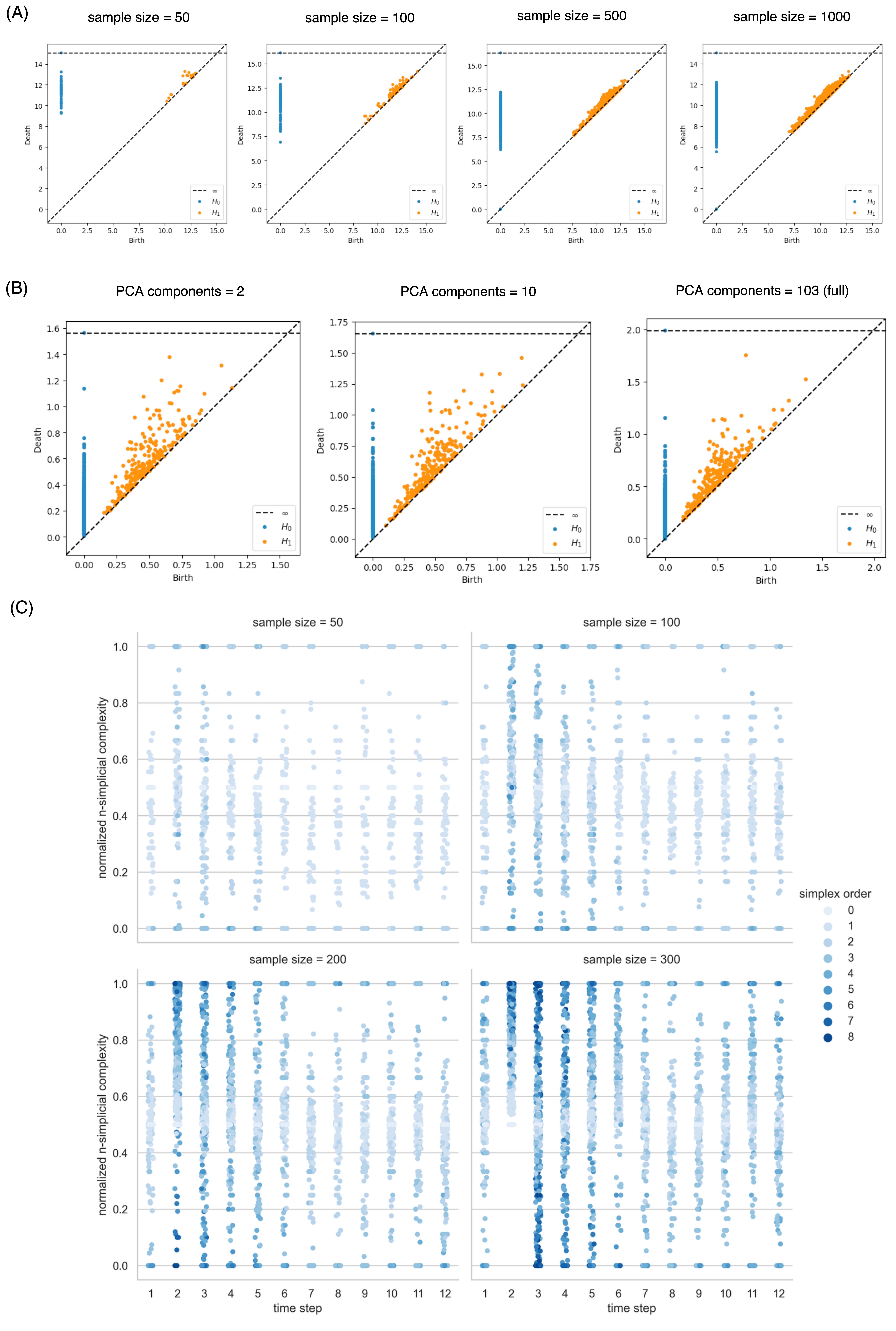} \hfill
\caption{\textbf{Sensitivity analysis} of the effective of witness sampling and PCA dimension reduction to the persistent homology and simplicial analysis. (A) Persistent diagrams of the dataset when sampling 50, 100, 500 and 1,000 data points. (B) Persistent diagrams when choosing the first 2, 10 and 103 (all dimensions) principal components. (C) The overall distribution of the normalized simplicial complexity doesn't change much when the sampling size at each time step arises from 10, 100, 200 to 300.}
\label{fig:sensitivity}
\end{figure}

We benchmark the scTSA method on the zebrafish single-cell RNA sequencing data with 38,731 cells, 25 cell types, over 12 time steps \cite{farrell2018single}. The dataset studies the embryogenesis, which is the process where the cells gradually differentiate into distinct fates through stages of transcriptional changes. The goal of this study is to facilitate a comprehensive identification of cell types with their time stamps in order to reconstruct their developmental trajectories (e.g. transcriptional states, branch points and asyncrhony). As the gene expression profiles they obtained from the vertebrate embryo are time stamped spanning 3.3–12 hours post-fertilization (hpf), it provides a perfect testbeds for time-series analysis to reconstruct transcriptional trajectories and characterize time-dependent development properties. 

We process the scRNA sequencing data into entries of 103 dimensions corresponding to the expression levels of 103 significant genes (which we select using scTDA). We then standardize the features by removing the mean and scaling to unit variance. Before we perform the persistent homology, we first embed the dataset into low-dimensional space using dimension reduction. In Figure \ref{fig:dr}, we embed the data using Principal Component Analysis (PCA), t-distributed Stochastic Neighbor Embedding (TSNE) \cite{van2008visualizing} and Uniform Manifold Approximation and Projection (UMAP) \cite{mcinnes2018umap} and color them by time steps. We observe that they all demonstrate a temporal gradient. 

We perform the persistent homology and temporal persistent homology on the scaled dataset. The $\tau$ in this case is set to be 1 (meaning that we only care about the linkage formed between consecutive time points). Figure \ref{fig:tph} compares the persistent diagram for the two approaches. From the persistent diagrams, the persistent features detected by the persistent homology are not noticeably different from the temporal persistent homology. While not a focus of this work, further study using downstream machine learning tasks can potentially pinpoint the benefits of these temporal persistent features.

For the simplicial analysis, we first group the data by their time steps. The data collected at the 12 time steps are highly imbalanced: 1 (2225 data points), 2 (200),  3 (1158),  4 (1467),  5 (5716), 6 (1026), 7 (4101), 8 (6178), 9 (5442), 10 (5200), 11 (1614) and 12 (4404). For each time points, we perform a witness sampling of 200 data points, since it is the lowest number of samples among all time points. We identifies the simplicial complexity to vary over the time, suggesting a potential better summary statistic with better distinction among time steps (Figure \ref{fig2}). The normalized simplicial complexity (computed as the ratio of the number of simplicial complexes discovered within the data over the number of those discovered within the null model) suggests an abundance of high-dimensional simplices over the null models. The existence of a significant number of high-dimensional simplices is observed for the first time in the single cell level. In all time points, the number of simplices of dimensions larger than 1 in the null model was far smaller than those found in the actual data. In addition, we observe this relative differences between what we discover in null models and the actual data increase drastically when the dimensions are higher. Furthermore, the number of low-dimensional simplices (up to dimension 3) of the data appears to be equal or smaller than the null models (with normalized complexity less than 1), suggesting a possible transfer from lower order clique structure to a higher-order structure. 

In order to investigate the tradeoff between the higher-order and the lower-order simplicial complexity in the developmental stages, we map the normalized 3-simplicial complexity against the normalized 1-simplicial complexity. Figure \ref{fig2C} suggests an a gradually increasing higher-order complexity starting from the 5th time point, and an overall below-null lower-order complexity in a monotonically increasing direction since the 2nd time point. Comparing to the null model, the presence of a much larger numbers of cliques across a range of dimensions in the single cell data suggests that the connectivity between these cells might be highly organized into numerous fundamental building blocks (e.g. proto-cell types) with increasing complexity.
These two figures both suggest that the gastrulation stage (time point 5 to 6) is a very critical stage in vertebrate development, matching the established understanding in the developmental biology that it is a process where the embryo begins the differentiation process to develop into different cell lineages \cite{gilbert2016developmental}. Before gastrulation, the embryo is a continuous epithelial sheet of cells. After the gastrulation stage, organogensis starts where individual organs develop within the newly formed germ layers.

This observation is further supported by the visualization of topological data analysis mapping. Figure \ref{fig3C} compares the network visualizations with and without the temporal filtration. We observe that, when color-labelled with the time points, the conventional topological data analysis outlines a progression of cellular development, but there are many subsequent time points in the middle of earlier timesteps. For instance, we see there are many dark blue nodes from the 11th or 12th time points in the middle of web where the majority of the nodes are earlier stages from the 5th to 7th. When using the temporal filtration (with $\tau$ set to be just 1 time step), we observe that the network has much more skeleton and branches, where each branching nodes consist only of points of the same time stamp. The gastrulation stage, which happens between the 5th and 6th time points, appears to belong to two separate tracks, supporting the hypothesis that after the notochord and prechordal plate territories become transcriptionally distinct, the gastrulation process refines the boundary between the two cellular populations \cite{farrell2018single}.

These filtrated simplicial architectures may also offer insights in cell lineage tracing. We perform the hierarchical clustering of the summary statistics computed from the transcriptome data of different cell types. We compare the result using the proposed normalized simplicial complexity versus the one using the Betti numbers (which is more conventionally used in many downstream topological data analyses). As shown in Figure \ref{fig:lineage}, the normalized simplicial complexity offers a more reasonable clustering performance as a more distinctive summary statistics than the Betti numbers by themselves.

\section{Discussion}
\label{sec:conclusion}

What is cellular complexity and what does the higher-order complexity mean? As an inquiry to this question, we explore the possibility of introducing the mathematical notion of higher-order simplicial complexes into analyzing distance-based single cell data. Benchmarked on a single cell gene expression data with multiple developmental stages, we propose the single-cell Topological Simplicial Analysis, and demonstrate that the simplicial complexity can be a well-defined summary statistic for celluar complexity. 

This investigation provides a scalable, parameter-free\footnote{By ``parameter-free'', we mean that it doesn't have arbitrary hyper-parameters that the users have to set in order to perform the analysis. The parameter $\tau$, instead, is a user-specified parameter that is relevant to the specific application and problem of interest. An analogy to a prediction model would be, the learning rate is an arbitrary hyper-parameter, and the prediction window would be a user-specified parameter relevant to the application.}, expressive and unambiguous mathematical framework to represent the cellular complexity with its underlying structure. Locally, these structures are characterized in terms of the simplicial complexes. Globally, these structures are characterized in terms of the cavities formed by these simplices. Topological cavities are usually formed and then later filled with the additions of new edges (and potentially, nodes). When computing the persistent homology, we perform a filtration process which innately tracks the formation and later filling of topological cavities of different dimensions. The temporal persistent homology characterizes the information of cavities with the lifespan of these topological objects. 
This framework reveals an intricate topology of cellular similarity which includes a vast number of cliques of cells and of the cavities that bind these cliques together. These topological summary statistics that captures the relationships among the high-dimensional cliques uncover the transcriptional differences in the connectivity of cells of different types during the graph reconstructions process.

From the scTSA visualization, we discover, for the first time in any single cell data, an abundant number and variety of higher-order cliques and cavities. Comparing to the control models, the framework measures a much higher number of high-dimensional cliques and cavities in the graph construction filtration process. The critical stage identified by the framework matches the current understanding in the developmental biology. Comparing with the statistics of Betti numbers, the normalized simplicial complexity demonstrates better distinctions between time points and cell types. 

Topological data analysis, like many other machine learning methods, have many empirical considerations related to sample sizes and dimensionality selections. 
To demonstrate the sensitivity of persistent homology to sampling size and reduced dimensions, we perform the following experiment. We use the full dimensions of the standard scaled dataset, vary the sampling size from 50, 100, 500 to 1,000 data points and compute their persistent diagrams. We then set the sample size to 1,000, vary the PCA dimensions to be the first 2, 10 and 103 (full) dimensions, and compute their persistent diagrams. We observe no clear difference. Then, we perform the simplicial analysis with witness sampling using sample size from 10, 100, 200 to 300. In this case, we observe a slightly higher numbers of higher order simplical complexes, but the overall shape and distinction between the time steps are maintained. Future study can investigate strategies of increasing the stability of the simplicial analysis to sample size.

In the introduction, we pose some open questions we wish to engage the field to discuss and investigate together, instead of answering them directly in this first work. Here we will briefly share our preliminary take on some specific ones:

Why does expression similarity deserve the name of complexity? To clarify, the expression similarity may not be a measure of complexity. However, the temporally connected higher order co-expression structure characterized by similarity can be a useful measure of complexity. If the task requires several agents to co-work together at the same time, or follow a specific sequence of actions by different agents, then it is more complex than a task which only requires a few agents or doesn't need to follow a specific sequence. The notion of similarity is usually related to clustering and thus, the separation of homogeneous groups. To extend on this understanding, the similarity relationships that are further constrained by temporal sequences would relate to functionally separating groups of homogeneous agents, and thus, potentially informative to their interactions. 

Is there a reason to believe gene expression similarity has something to do with interactions rather than reflecting the number of similar cells that happen to be present in the sample? These temporally constrained gene expression similarity can both reflect the number of similar cells that coexist at the same time, but also potentially related to the some level of functional interactions, as discussed above. We wish to leave further investigations on what type of interactions for future work, and welcome discussions and critique in these interpretations.

Finally, there are other potentially applicable questions we can explore: Can we determine developmental stages without physiological features? Can we generate pseudo-time series based on single cell sequencing data? And most importantly, does the vast presence of high-dimensional cliques suggest that the interaction between these cells is organized into fundamental building blocks of increasing complexity? Through this inquiry with topological simplicial analysis, we can form such hypothesis that the cells organize themselves into high-dimensional cliques for certain functional or developmental reasons. Further research includes developing mechanistic theories behind the emergence of such high-dimensional cellular cliques and experimentally testing these hypotheses to reveal the missing link between functions and cellular complexity.

\section{Conclusions}

In summary, our work describes a novel scalable and unsupervised machine learning\footnote{By ``machine learning'', we refer to the general goal of building a model that learns from the data. The topological data analysis is a class of unsupervised learning method. The topological features identified from the process can be further applied to downstream machine learning tasks, such as the hierarchical clustering of cellular lineage.} method that facilitate the understanding and solutions three main technical challenges in bioinformatics:

\subsection{A lack of time-series analytical methods in quantifying the underlying temporal skeleton within the manifold of the similarities among data points}

In persistent homology and mapper visualization, our temporal filtration uses a user-specified time separation parameter $\tau$, which can be either discrete (consecutive time steps) or continuous (by a time delay quantity). This enables the computation of persistent components that is computed only on data points that are temporally proximal, and thus, provides a temporal skeleton representation. In the simplicial analysis, we can group the data points by time steps, and compute the normalized simplicial complexity as a quantity to inform the ecology of cells in the transcriptomic feature space.

\subsection{A lack of scalable computational methods to characterize single-cell sequence signals in the scale of 10k+ data points, while the single-cell sequencing data are dominating bioinformatics in recent few years}

The usage of witness sampling and dimension reduction enable the computation of persistent homology to large numbers of high-dimensional data points. Sampling is also a required step to compare the topological features in groups of data points with different count numbers. The normalization against null distribution of the data sample partly corrects for the amplification effect of higher-order topological quantities. The usage of dimension reduction techniques such as PCA help with data management and computation without a significant loss of performance.

\subsection{A lack of insight and interpretation that connects the mathematical language of algebraic topology to the physical references to the biological phenomena}

In the introduction and discussion, we initiate the discussion of the interpretations of the topological properties. More specifically, we point out how the temporally directed relationships among data points can be related to functionally separating groups of homogeneous agents in the feature space, and thus,  potentially informative to their interactions. With our temporal-directed treatment of filtration or grouping techniques, our study is a small but first step to use topological data analysis as not only a descriptor tool for static manifold, but also, in the future, a discovery tool of dynamic or mechanistic components. Our goal in this work is not to fully answer the question of interpreting the biological insights topological properties, but to further motivate and facilitate our understanding to the question. As more techniques of topological data analysis are applying to biological problems, we wish to encourage the discussion and critique from the biology and machine learning research community. 

\subsection{Summary}

In summary, we propose a new family of filtrations for longitudinal time-series multidimensional data along with auxiliary data analysis tools. We demonstrate our application to the temporal inference problems using a set of time-resolved gene expression data. The key technique, called \textit{temporal filtration}, substitutes a conjunctive distance and time threshold for the conventional distance threshold for point cloud data augmented with time stamps. In addition to persistent homology, mapper constructions, and the use of witness sampling with this technique, an original set of standardized summary statistics, the \textit{normalized simplicial complexities}, are proposed. These techniques are used to conduct an exploratory analysis of zebrafish embryonic development through the lens of longitudinal single-cell RNA sequencing data. The applications showcase clear improvements in the interpretability of visualizations compared with a cross-sectional approach and suggest that the key events in the evolution of a biological system can be more effectively detected using normalized simplicial complexity than using Betti numbers. Other than the biological application in single-cell genomics, the time-series problem is especially a topic that is applicable beyond the application proposed in our work, and thus a major interest in the unsupervised machine learning communities dealing with high-dimensional time series signals.

\bibliographystyle{apacite}
\bibliography{main}  






\end{document}